\documentclass[letterpaper]{article} %, draft

\usepackage{natbib,alifeconf}  %% The order is important
\usepackage{algorithm}
\usepackage{algpseudocode}
\newlength\myindent
\newcommand\mydots{\makebox[1em][c]{.\hfil.\hfil.}}

\usepackage[export]{adjustbox}[2011/08/13]
\usepackage{hyperref}
\hypersetup{hidelinks,
colorlinks=true,linkcolor=ourblue,
urlcolor=ourblue,citecolor=ourviolet,
bookmarksopen=false}

\usepackage{csquotes}
\usepackage{comment}
\usepackage{xcolor}
\usepackage{ifthen}
\usepackage[super]{nth}
\usepackage{amsfonts}
\usepackage{epigraph}
\usepackage{csquotes}

\let\oldquote\quote
\let\endoldquote\endquote
\renewenvironment{quote}[2][]
  {\if\relax\detokenize{#1}\relax
     \def\quoteauthor{#2}%
   \else
     \def\quoteauthor{#2~---~#1}%
   \fi
   \oldquote}
  {\par\nobreak\smallskip\hfill--- \quoteauthor%
   \endoldquote\addvspace{\bigskipamount}}

\PassOptionsToPackage{fleqn}{amsmath} % Math environments and more by the AMS
\usepackage{amsmath}
\usepackage{comment}

\usepackage{xspace}
\makeatletter
\DeclareRobustCommand\onedot{\futurelet\@let@token\@onedot}
\def\@onedot{\ifx\@let@token.\else.\null\fi\xspace}

\makeatother

\usepackage{xcolor}
\definecolor{ourblue}{rgb}{0.368,0.507,0.71}
\definecolor{ourgreen}{rgb}{0.56,0.692,0.195}
\definecolor{ourred}{rgb}{0.923,0.386,0.209}
\definecolor{ourviolet}{RGB}{59, 58, 126}
\definecolor{ourorange}{RGB}{232, 122, 18}
\newcommand{\fs}[1]{{#1}}
\title{The dynamical regime and its importance for evolvability, task performance and generalization.}

\author{Jan Prosi$^{1, 2}$, Sina Khajehabdollahi$^{1, 3}$, Emmanouil Giannakakis$^{1, 2}$, Georg Martius$^{3}$ \and Anna Levina $^{1, 2, 4}$ \\
\mbox{}\\
$^1$ Department of Computer Science, University of T\"ubingen, T\"ubingen, Germany\\
$^2$ Max Planck Institute for Biological Cybernetics, T\"ubingen, Germany\\
$^3$ Max Planck Institute for Intelligent Systems, T\"ubingen, Germany \\
$^4$ Bernstein Center for Computational Neuroscience  T\"ubingen, T\"ubingen, Germany}

\begin{document}
\maketitle

\begin{abstract}
It has long been hypothesized that operating close to the critical state is beneficial for natural and artificial systems.
We test this hypothesis by evolving foraging agents controlled by neural networks that can change the system's dynamical regime throughout evolution.
Surprisingly, we find that all populations, regardless of their initial regime, evolve to be subcritical in simple tasks and even strongly subcritical populations can reach comparable performance.
We hypothesize that the moderately subcritical regime combines the benefits of generalizability and adaptability brought by closeness to criticality with the stability of the dynamics characteristic for subcritical systems.
By a resilience analysis, we find
%To test this, we perturb the genome and life-conditions of fully evolved agents and check their resilience.
%We find
that initially critical agents maintain their fitness level even under environmental changes and degrade slowly with increasing perturbation strength.
On the other hand, subcritical agents originally evolved to the same fitness, were often rendered utterly inadequate and degraded faster. % with genetic perturbations.
We conclude that although the subcritical regime is preferable for a simple task, the optimal deviation from criticality depends on the task difficulty:
for harder tasks, agents evolve closer to criticality.
%the harder the task, the more towards criticality agents will evolve.
Furthermore, subcritical populations cannot find the path to decrease their distance to criticality.
In summary, our study suggests that initializing models near criticality is important to find an optimal and flexible solution.
\end{abstract}

\section{Introduction}
Operating close to the critical point at a phase transition between order (subcritical) and disorder (supercitical) has long been associated with optimal performance of complex systems. Several biological systems, such as gene regulatory networks \citep{balleza2008critical, ramo2006perturbation}, neural networks \citep{tkacik2015thermodynamics, schneidman2006neural_cultures}, collectively behaving cells \citep{halley2009collective_cells, depalo2017collective_cells} or swarms \citep{cavagna2010scale, chate2014insect} have been shown to operate close to a critical point.
 Criticality has been associated with an ability to solve complex tasks \citep{villegas2016noise_subcritical},  flexibility towards changes in the environment and good evolvability \citep{aldana2007robustness_evolvability} in complex living systems \citep{kauffman1993origins}.
 All these properties provide an adaptive advantage in natural environments, leading to the assumption that evolutionary dynamics push living systems close to the critical regime.
 \par
 \begin{figure}
\centering
%width 2.3in
\includegraphics[width=.95\columnwidth]{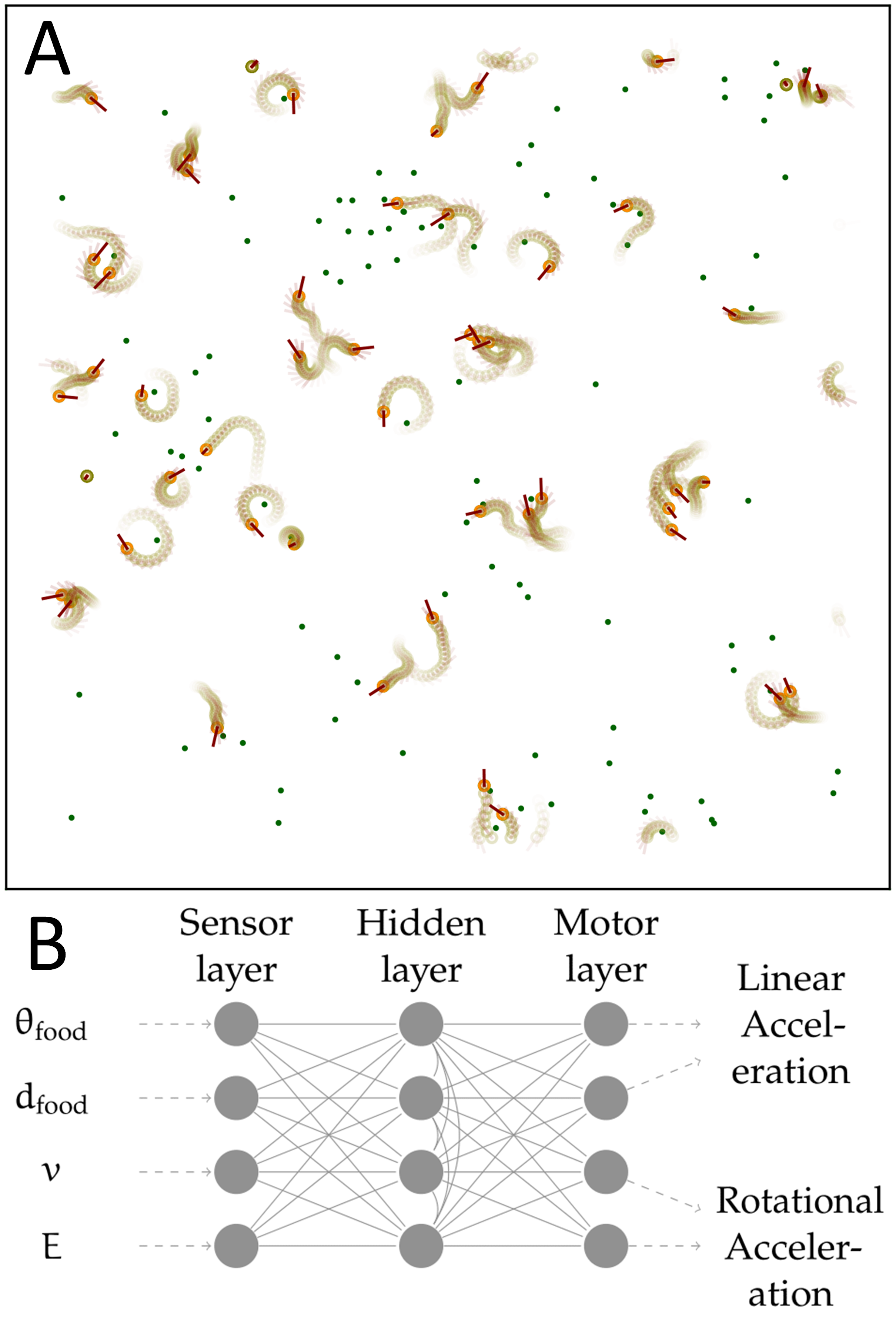}
%\includegraphics[width=3.4in]{Figures/Methods_figure_cut.png} %original width
%\caption{$2$D environment and network topology}
\caption{
Snapshot in time of population dynamics and schematic representation of the control network. \textbf{A:}
%$2$D environment with periodic boundary conditions, where the organisms are collectively evaluated for their ability to collect energy in a foraging task. An organism can gain energy by running over the green food particles.
Environment with 50 organisms (red circles with trails) foraging for food (green dots).
\textbf{B:} A network with $12$ neurons. Four sensory, four hidden and four motor neurons. All potentially allowed edges are displayed. The exact topology and edge weights are subject to evolution by the EA.
}
\label{fig:methods}
\end{figure}

 On the other hand, it has been suggested that the ubiquitous presence of noise in nature pushes living systems into a more robust subcritical regime.
 For example, in an evolutionary model of Random Boolean Networks (RBNs) decreasing the system size, making the task less complex, or introducing noise to the system pushes the optimal regime further into the subcritical range \citep{villegas2016noise_subcritical}. Similarly \citet{pauli_ramo2007_critical_information_propagation_subcritical_noise} observed, that information propagation is maximized in critical RBNs, however the optimal regime shifts slightly into the subcritical regime under the presence of noise.
 A related phenomenon has been observed on neuromorphic chips, where simpler tasks lead to optimal behavior in the subcritical regime, whereas harder tasks require progressively more critical dynamics \citep{cramer_control_2020}.
 Finally, for some applications, the combination of systems at different distances to criticality is shown to lead to optimal results \citep{zierenberg2020tailored}. The supercitical state has been universally observed to perform poorly \citep{villegas2016noise_subcritical, kauffman1993origins}.

The benefits of criticality for the evolvability of living systems have been associated with the genotype-phenotype coupling.
Specifically, it has been shown  \citep{dejong2006evolutionary_unified_book} that a tight genotype-phenotype coupling leads to optimal evolvability.
Due to this coupling, the dynamical regime has an impact on the properties of the fitness landscape. In an RBN model, the super- and subcritical regimes were shown to disturb the genotype-phenotype coupling \citep{kauffman1993origins} and lead to either very rugged or overly flat fitness landscapes.
A rugged fitness landscape means that the evolutionary dynamics are just a random search -- inefficient in high dimensions~\citep{kauffman1987waiting_times_double}, whereas
a very flat landscape strongly dampens the optimization process. Both phenomena lead to a complexity catastrophe, where an increase of system size leads to a failure to discover satisfying solutions with evolutionary search.
Critical RBNs result in an intermediately rugged fitness landscape which allows for efficient hill climbing search and is less prone to the complexity catastrophe.

We study how the dynamical regime of populations of evolving organisms influences their ability to solve a task. Our investigation is conducted in a simple foraging game of scalable difficulty.
This system allows us to analyze the changes in the dynamical regimes during the evolution.
In addition, we propose a potential answer to the question of which dynamical regime demonstrates the best performance and stability with respect to changes in the environment.

\section{Methods}

\begin{figure*}
\centering
%width 2.3in
\includegraphics[width=\textwidth]{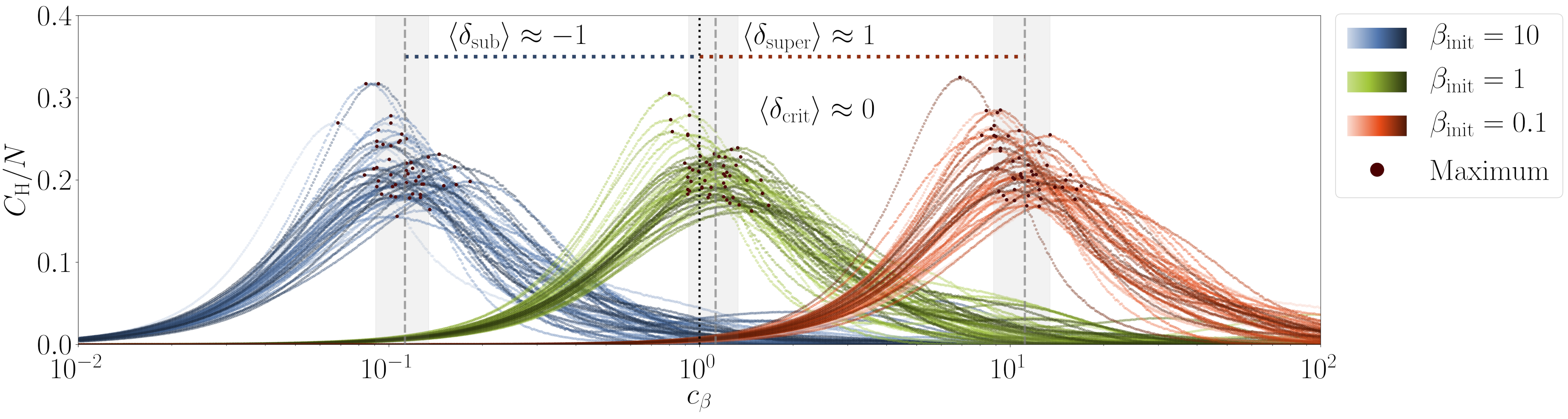}\vspace{-.8em}
\caption{The dynamical regime of a network can be calculated by scaling the inverse temperature with a factor $c_\beta$ to measure its heat capacity and measuring its distance to the corresponding peak.
Heat capacity (Eq.~\ref{eq:Heat_Cap}) of the Ising  networks (Figure~\ref{fig:methods}) for 50 initially subcritical ($\beta_{\mathrm{init}} = 10$, blue), critical ($\beta_{\mathrm{init}} = 1$, green), and supercritical ($\beta_{\mathrm{init}} = 0.1$, red) organisms as a function of $c_\beta$.
For each organism it reaches the maximum (marked by a dot) at individual values $c_\beta = c_\beta^\mathrm{crit}$.
Dynamical regime $\delta = \log (c_\beta^\mathrm{crit})\approx - \log (\beta_{\mathrm{init}}).$
%The parameter $\delta$ characterizes the dynamical regime of an organism. It describes the deviation of the maximum at $\beta_\mathrm{fac}^\mathrm{crit}$ to the original simulation condition at $\beta_\mathrm{fac}=1$ on a logarithmic scale.
The displayed populations are unevolved and the resulting dynamical regimes closely correspond to their respective $\beta_{\mathrm{init}}$.}
\label{fig:heat_cap}
\end{figure*}

\fs{We investigate a 2D environment where organisms controlled by individual neural networks forage for food}.
Each organism gains energy by eating food particles and consumes energy by moving. The organisms eat the food particles as soon as they run over them.
We can increase the difficulty of the task by requiring the organism's velocity to be below a certain threshold when running over food in order to be able to consume it.
The fitness of an organism is determined by its average energy throughout its lifetime, and an evolutionary algorithm (EA) optimizes the network that controls the organism to maximize its fitness.

\subsection{Organism}

The organisms in our model are controlled by an Ising neural network (INN) that has been previously used by \citet{aguilera2017ising_neural_network} as well as \citet{khajehabdollahi2020sinas_paper}.
The Ising network consists of $N$ neurons that can be in one of two states $ s_i \in \{-1, 1 \}, i = 1, \dots N$. There are sensory neurons that only receive input from the sensors and motor neurons that control the agent. All other neurons are hidden units.
The connectivity of the network is described by the weight matrix $J \in [-2, 2]^{N \times N}$
and the adjacency matrix $A \in \{0,1\}^{N \times N}$, as shown in Figure~\ref{fig:methods}B.
The adjacency matrix is such that no direct connections between sensor and motor neurons exist.
Following the Ising model, each network state has an associated energy
\begin{eqnarray}
E(s_1,\ldots,s_N)= -\sum\limits_{\{i,j \mid A_{ij}=1\}} J_{i,j} s_i s_j
\label{eq:ising_net_energy}
\end{eqnarray}
where $\{i, j \mid A_{ij}=1\}$ is the set of connections.
The network stochastically minimizes the energy by following Glauber dynamics:
%We repeatedly iterate through each non-sensor neuron of the network in a random order.
At each network iteration, all non-sensor neurons are updated in a random order and the state of neuron $i$ changes from $s_i$ to $-s_i$ with probability:
\begin{eqnarray}
p_i &=& \frac{1}{1+e^{\beta \cdot \Delta E_i}},
    \label{eq:spin_flip}\\
\Delta E_i &=& E (s_1,\mydots,s_i,\mydots,s_N) - E (s_1,\mydots,-s_i,\mydots,s_N), \nonumber
\end{eqnarray}
where $\beta$ is the inverse temperature of the network ($\beta = 1 / (T \cdot k_B)$,  $k_B$ is the Boltzmann constant which we set to one and omit for simplicity) and $\Delta E_i$ is the change in the energy of the network that is caused by the spin-flip of the $i_{th}$ neuron (changing its state $s_i$ to $-s_i$).
The energy change $\Delta E_i$ is determined by the connectivity matrix $J$ and the states of neighboring neurons.
A negative energy change leads to a greater likelihood of a flip. The parameter $\beta$ controls the likelihood of energetically unfavorable flips.
A larger $\beta$ leads to deterministic network behavior dominated by the connectivity, whereas a smaller $\beta$ leads to more random behavior.
In principle, many iterations would be required to converge to  equilibrium. For practical reasons, we perform a fixed number of iterations (10) to process a new sensor input and create motor commands.

% The INN inheres a phase transition between the deterministic subcritical phase and a random supercritical phase.

%\subsection{Environment}
An organism has four input neurons that receive information about the angle $\theta_\mathrm{food}$ and distance $d_\mathrm{food}$ from the closest food particle as well as its own velocity $v$ and energy $E$.
Moreover, each organism has four output neurons that control linear and rotational acceleration (2 neurons each) and $N_h$ hidden neurons (Figure~\ref{fig:methods}B).
For most simulations we take $N_h = 4$, to test scalability of learning we also consider a network with $N_h = 20$.
At the beginning of each simulation, an organism is provided with an amount of initial energy $E^\mathrm{init}=2$.
Movement reduces energy and consuming food particles increases it.
We consider two versions of this environment:  In the \emph{simple task}
organisms consume food when passing over it.
In the \emph{hard task} organisms have to slow down and almost stop to be able to consume food.
Unless stated differently, a simulation lasts for a \emph{lifetime} of $t=2000$ time steps after which the evolutionary algorithm (EA) is applied, and the task is simple.
50 INN-controlled organisms are placed in a $2$D environment with periodic boundaries and ever-respawning food particles, conserved to a value of 100 (Figure \ref{fig:methods}A).

\subsection{Evolutionary algorithm}
The evolutionary algorithm applied to the INNs consists of a combination of elitism, mutation, and mating.
At the end of the 2D simulation described above, the fitness of each organism is defined as their mean energy throughout their lifetime.
Subsequently, the 20 fittest organisms are selected for reproduction.

The next generation of organisms is produced by applying a combination of copying, mutation, and mating procedures to the selected individuals.
The copying algorithm transfers some of the fittest individuals into the next generation unchanged.
The mutation algorithm adds or deletes edges in $A$ (connections not present in Figure~\ref{fig:methods}B cannot be added), re-samples a random edge weight in $J$ from a uniform distribution $\mathcal{U}(-2, 2)$, and perturbs the inverse temperature with multiplicative noise $\beta' = \beta \cdot \Delta \beta$ where $\Delta \beta \sim \mathcal{N}(1,\,0.02)$.
Finally, the  mating algorithm takes a weighted average of the connectivity $J$ and inverse temperature $\beta$ from two parents to produce an offspring.
In most of our simulations, the EA iterates for 4000 generations.

\begin{figure*}\centering
\begin{tabular}{ c c c }
%\numexpr3*(1677/1770)\relax
\textbf{A.} Simple Task & \textbf{B.} Simple Task & \textbf{C.} Hard Task\\
$12$ Neurons & $28$ Neurons & $12$ Neurons \\
 \includegraphics[height=3.5cm]{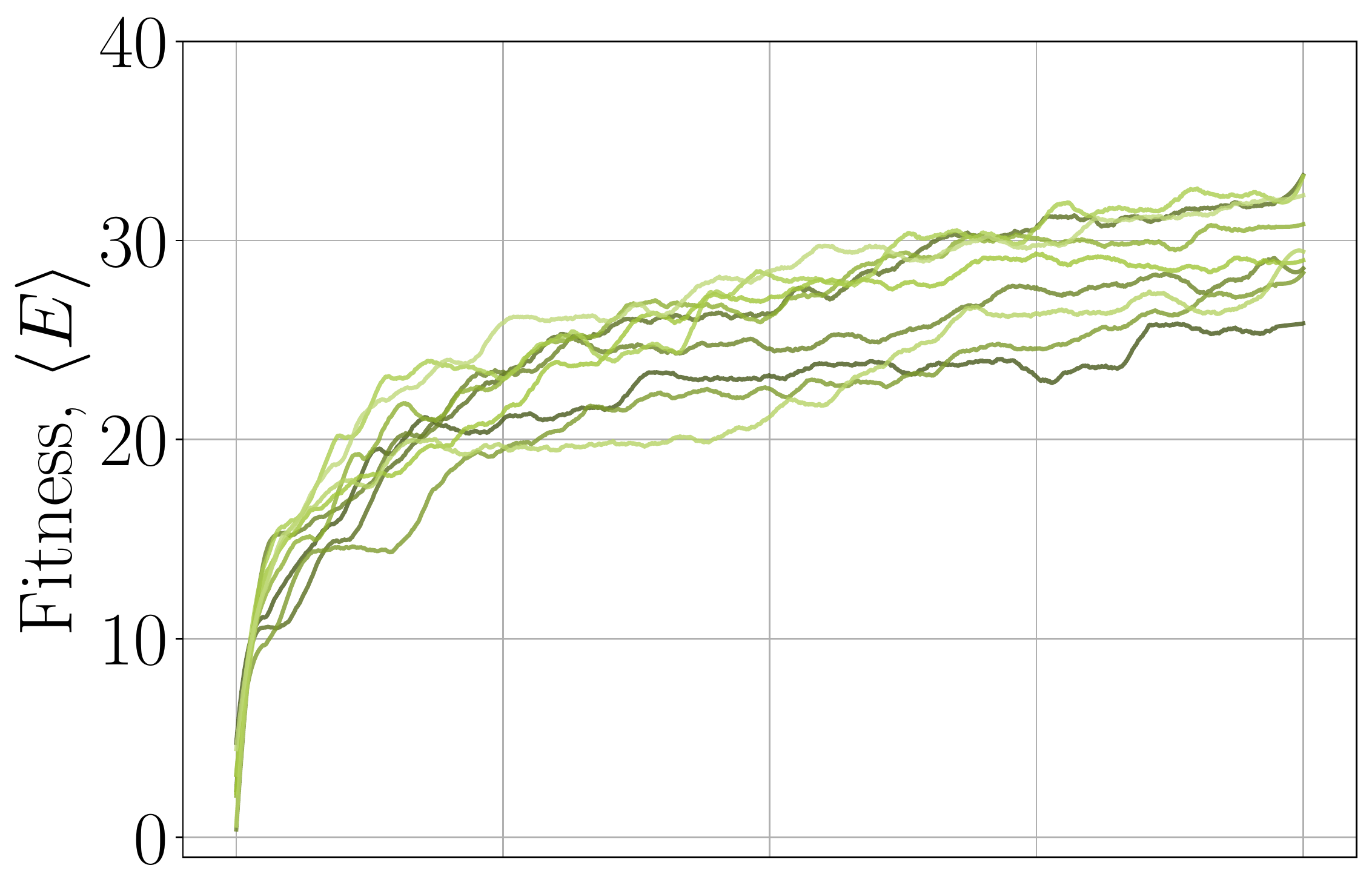} & \includegraphics[height=3.5cm]{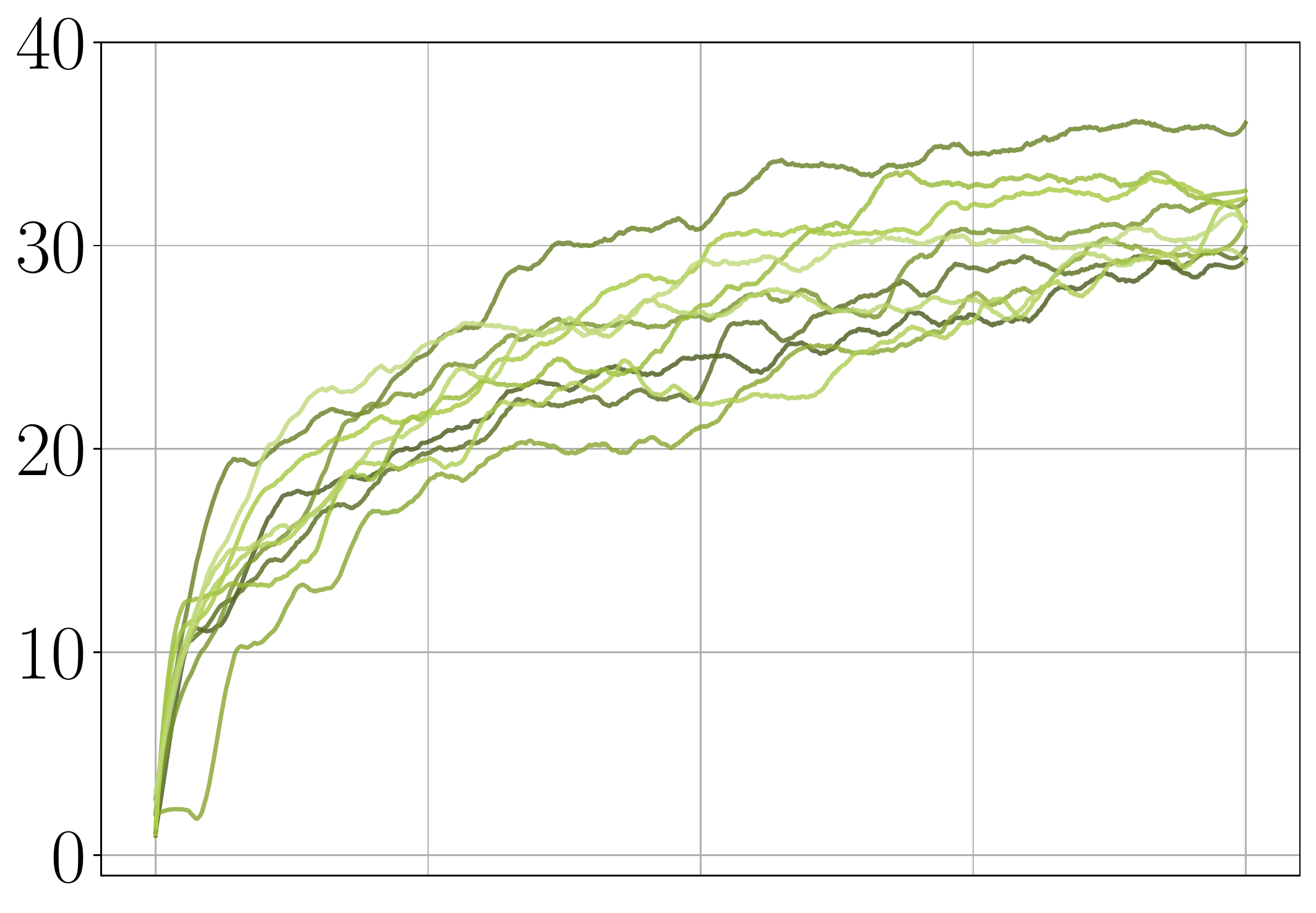} & \includegraphics[height=3.5cm]{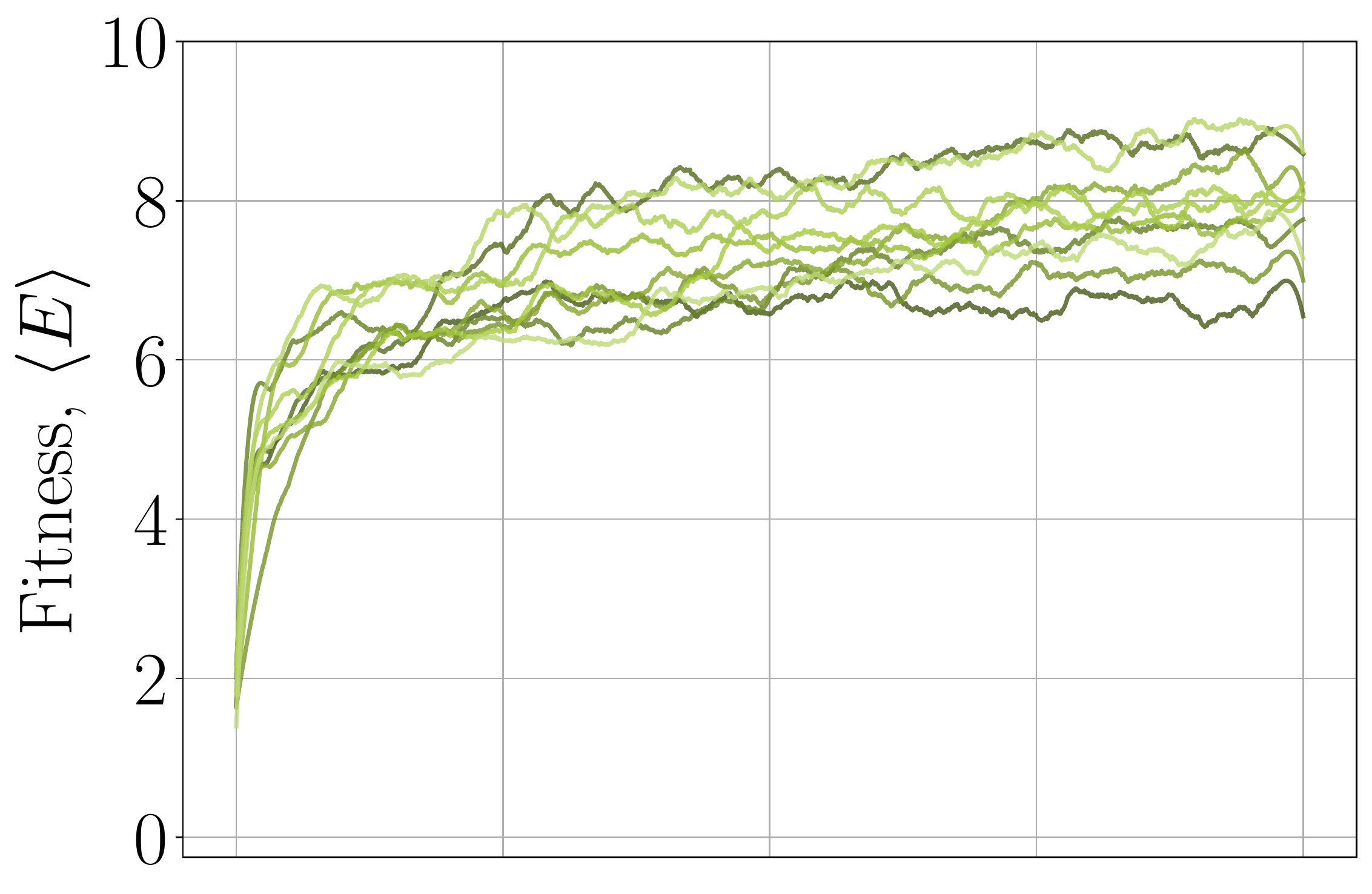} \\
 \includegraphics[height=4.0338cm]{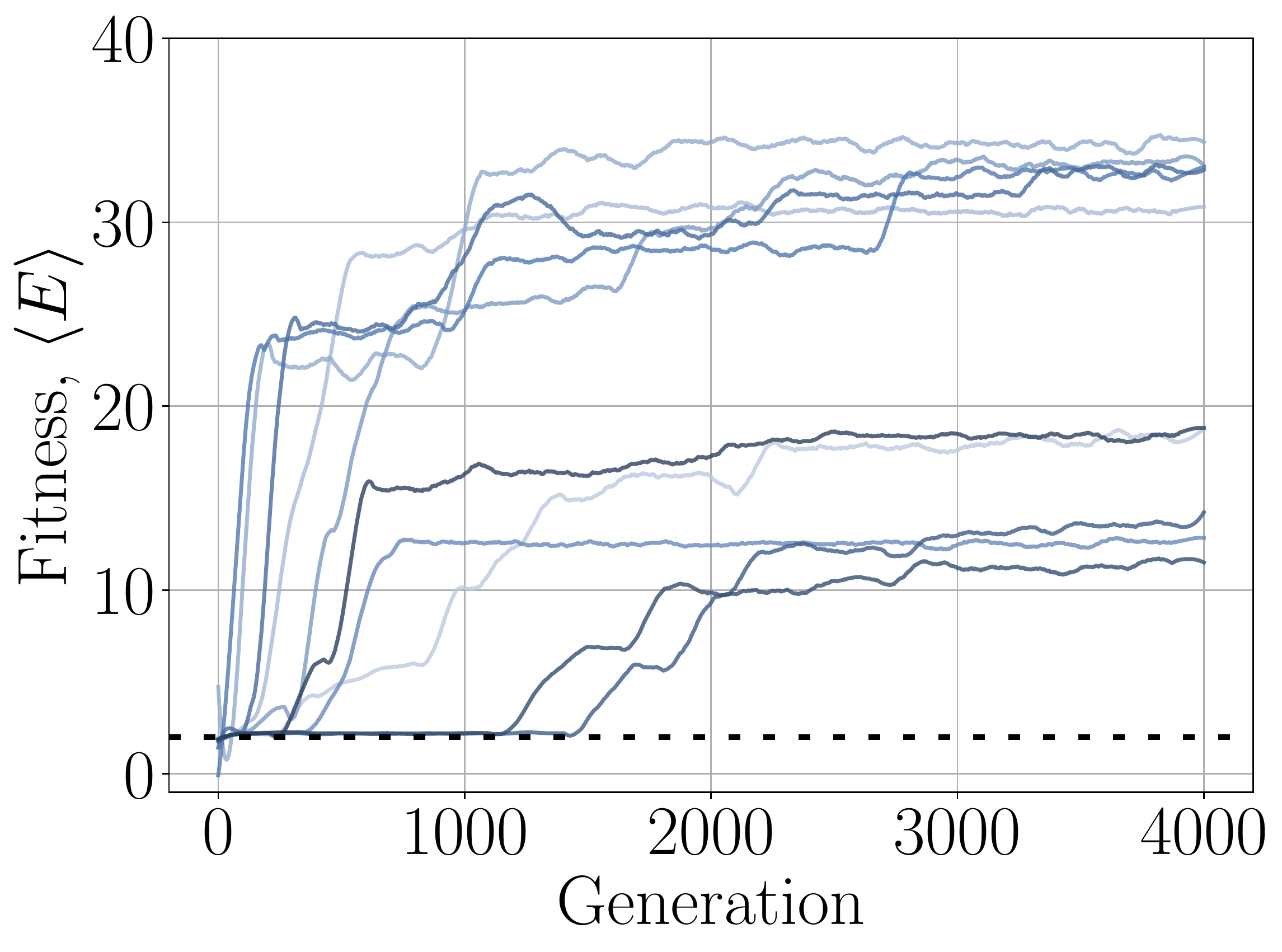} & \includegraphics[height=4.0338cm]{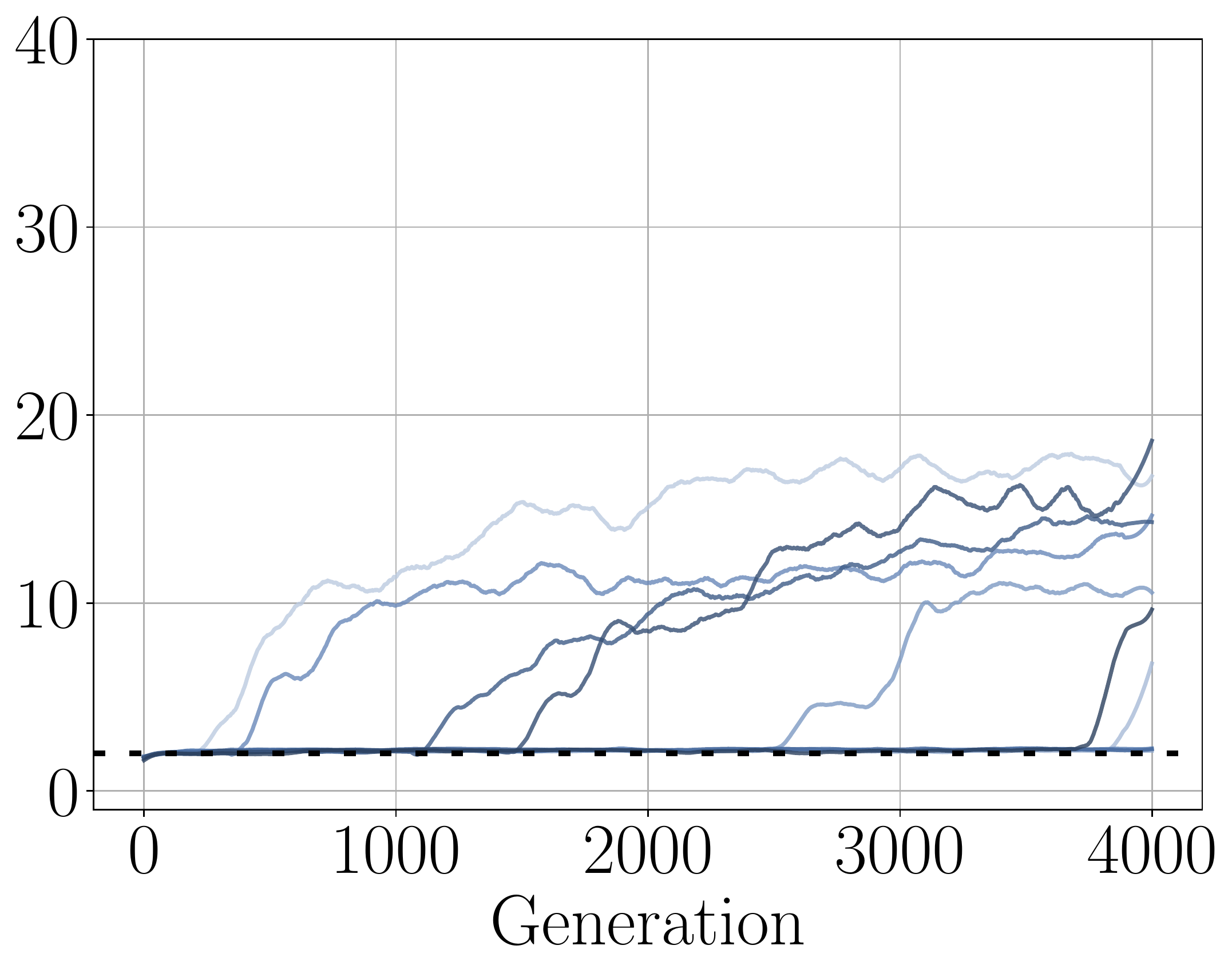} & \includegraphics[height=4.0338cm]{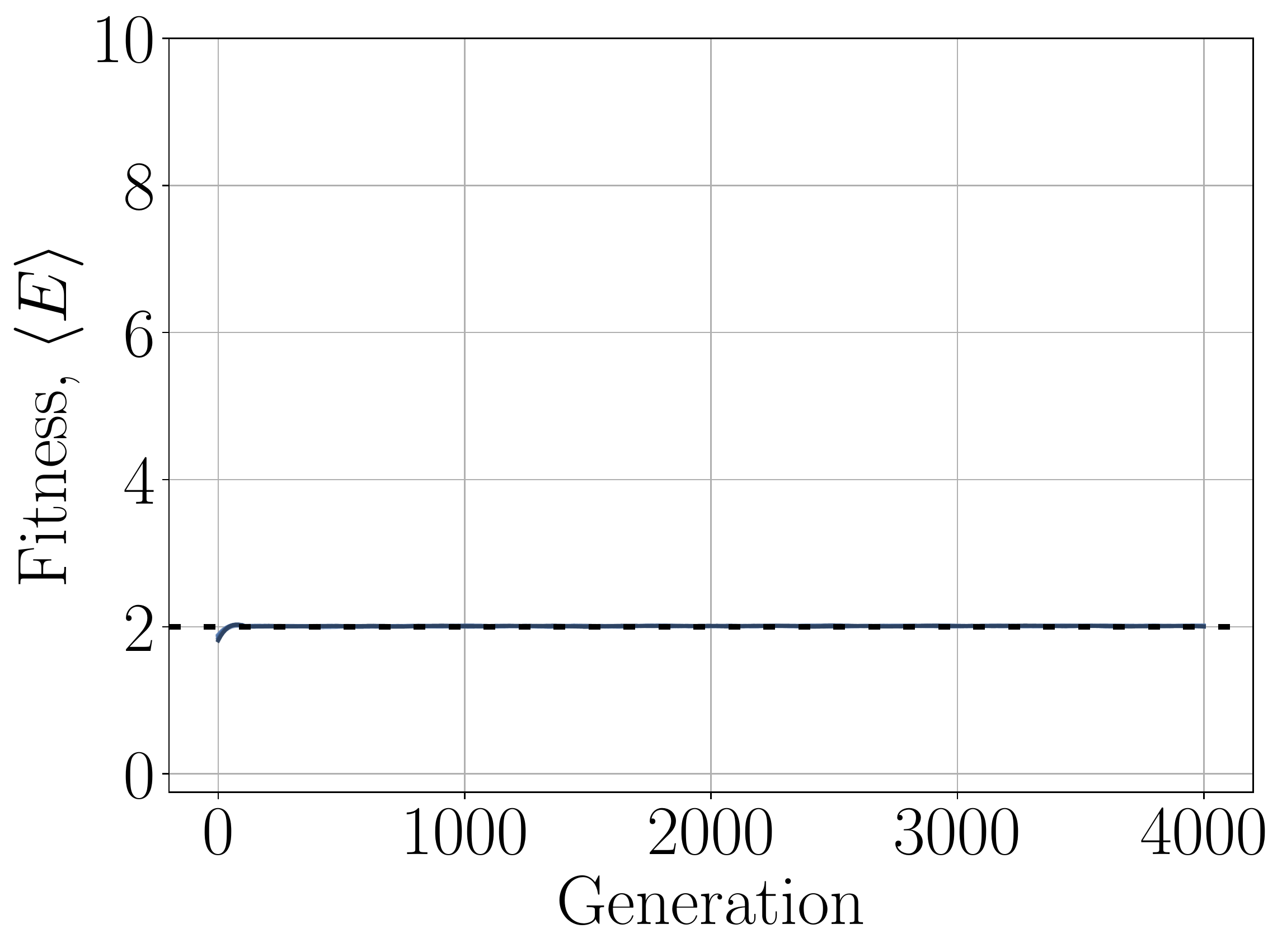}
 % Lower figures height has to be scaled up by factor (2040 / 1770)
\end{tabular}\vspace{-.5em}
\caption{Critically initialized populations can be successfully evolved in different circumstances, whereas for subcritically initiated populations, a harder task or an increased system size can lead to the breakdown of evolutionary dynamics.
For each panel 10 initially critical ($\beta = 1$, green, top row) or initially subcritical ($\beta = 0.1$, blue, bottom row) populations evolve for 4000 generations.
The dashed horizontal line at fitness = 2 in the subcritical panels correspond to the organisms' initial energy. It can be seen in the 12 neuron, hard task panel (bottom right) that the network is unable to achieve a fitness above its starting value.}
\label{fig:fitness_plots_vs_generation}
\end{figure*}

\subsection{Defining the dynamical regime of an organism}

\renewcommand{\thealgorithm}{}
\begin{algorithm}[tbp]
\caption{}
\label{alg:model}
\begin{algorithmic}[1]

%\Procedure{MyProcedure}
%\State $\text{population} = \text{initialize population}(\text{population size})$
\State {\color{ourred} \textbf{Evolutionary Algorithm}}
\Comment{\textit{Evolves $J$, $A$, and $\beta$}}
\For{$\text{generation} = 1 \textbf{ to } \text{Generations}$}
\State {\color{ourred} \textbf{Foraging Game in $2$D environment}}
\hspace{\algorithmicindent}\For{$t = 1 \textbf{ to } \text{Time Steps}$}
%\State $\text{Update organism's sensor neurons}$
\State $\text{Update Sensor Neurons}(\text{organism})$
\State {\color{ourred} \textbf{Update INN}} %\Comment{\textit{Metropolis algorithm alters neuron states towards energy minimum}} %\parbox[t]{.35\linewidth} : zahl vor linewidth muss auf platz angepasst werden!
\hspace{\algorithmicindent}\hspace{\algorithmicindent}\For{$\text{iter} = 1 \textbf{ to } \text{Network Iterations}$}
\hspace{\algorithmicindent}\hspace{\algorithmicindent}\hspace{\algorithmicindent}\For{$\text{non-sensor neuron} \textbf{ in } \text{INN}$}
\State $\text{Potential spin-flip(neuron)}$ %Glauber Step
\EndFor
\EndFor
\State $\text{Read Motor Neurons}(\text{organism})$
\State $\text{Move in 2D environment}(\text{organism})$
\EndFor
\State $\text{Evolve}(\text{population of organisms})$
\State $\text{Reset 2D environment}$
\EndFor
\end{algorithmic}
\end{algorithm}

%\fs{The dynamical regime of an organism (sub-, super-, critical) is determined by the  divergence point of the heat capacity $C_H$ (Equation \ref{eq:Heat_Cap}).}

We use the heat capacity from statistical physics to derive a measure of an organism's dynamical regime (sub-, super-, critical).
In our finite system, we estimate the putative divergence point by changing the inverse temperature $\beta$ multiplying it with a scaling constant $c_\beta$.
This change of temperature influences how likely the state of the neurons will flip (Eq.~\ref{eq:spin_flip}), and thus change the equilibrium distribution of energies $E$ (Eq.~\ref{eq:ising_net_energy}). Now we search for a $c_\beta$ that delivers the maximal value of the heat capacity $C_H(c_{\beta})$, defined as:
\begin{eqnarray}
    \label{eq:Heat_Cap}
    %C_o(\beta_{\mathrm{fac}}) = \beta_{\mathrm{fac}} \cdot \beta_o (\langle E^{net^2}_o \rangle - \langle E^{\mathrm{net}}_o \rangle ^2 )
    C_H(c_{\beta}) = \frac{1}{T^2} \textrm{Var}(E) = c_{\beta}^2  \beta^2  \textrm{Var}(E).
\end{eqnarray}
We define the $c_\beta^{\mathrm{crit}} = \underset{c_{\beta}}{\mathrm{argmax}}\, C_H(c_{\beta}) $. An analogous procedure was used in~\citep{tkacik2015thermodynamics}.

We define the distance of the network from the critical point by the logarithm of the scaling factor required to bring the network to criticality, $\delta = \log(c_\beta^{\mathrm{crit}})$.
For unevolved organisms (first generation, Figure \ref{fig:heat_cap}), the relationship between $\beta_{\mathrm{init}}$ and $\delta$ can be approximated by:
%The logarithm $\delta = \log(c_\beta^{\mathrm{crit}})$ indicates the distance of a networks from its critical point (since $c_\beta^{\mathrm{crit}}$ is the necessary amount of scaling required to push the network into criticality). We observe that only for an unevolved  (Generation $= 0$), the relationship between $\beta$ and $\delta$ can be approximated by Equation \ref{eq:beta_delta_approx}.
\begin{eqnarray}
\delta \approx - \log \beta_{\mathrm{init}}, \: \beta_{\mathrm{init}} \in [-0.1, 10].
\label{eq:beta_delta_approx}
\end{eqnarray}
On a technical side, for the stability of the numerical procedure for finding the equilibrium distribution to estimate $C_H(c_{\beta})$ we initiate the motor and hidden neurons in the state $(s_1,\ldots,s_N)$ delivering minimum of $E(s_1,\ldots,s_N)$, while keeping sensory inputs fixed to the observed values.
During the evolution, the distance from the critical point can (and will) change.

% \figdetail{In all Figures except for Figure 2 and Figure 7B, we are only using 20 most fit organisms to calculate the means}

%Due to the negative real-valued edge weights and fixed sensor neurons, the exhaustive thermalization gets stuck in local energy minima in the subcritical regime for our Ising implementation. A sudden drop into lower minima can lead to an increased variance of $C_H$ and thus outliers. We solved this by initializing the network at its lowest energy state prior to exhaustive thermalization. $C_H$ (Equation \ref{eq:Heat_Cap}).

\section{Results}

\subsection{Convergence of evolution}
Populations of different initial states follow distinct evolutionary strategies but are all able to solve the standard foraging task.
We observe evolution for 4000 generations, concentrating on the populations initiated at the supercritical ($\beta = 10$, $\delta \approx 1$), critical  ($\beta = 1$, $\delta \approx 0$) and subcritical ($\beta = 0.1$, $\delta \approx -1$) regimes.
Critical populations begin to rapidly gain fitness from the first generation in all EA realizations (Figure~\ref{fig:fitness_plots_vs_generation}A).
The gradual and stable increase of fitness of the initially critical population suggests that successful hill climbing of the fitness landscape is taking place.
In contrast, for subcritical populations, fitness mainly evolves via random jumps, and only half of the simulations reach the same fitness as the critical populations after 4000 generations (Figure~\ref{fig:fitness_plots_vs_generation}A).
Such fitness dynamics indicate a random search strategy, which often leads to a population getting trapped in local maxima for extended periods of time.
Confirming the previous observations by~\citep{khajehabdollahi2020sinas_paper}, we see that supercritical populations, after an initial random period, follow the same path as the critical ones.

\begin{figure*}\centering
\begin{tabular}{ c c }
%\numexpr3*(1677/1770)\relax
\textbf{A.} Simple Task & \textbf{B.} Hard Task\\
 \includegraphics[height=1.8in]{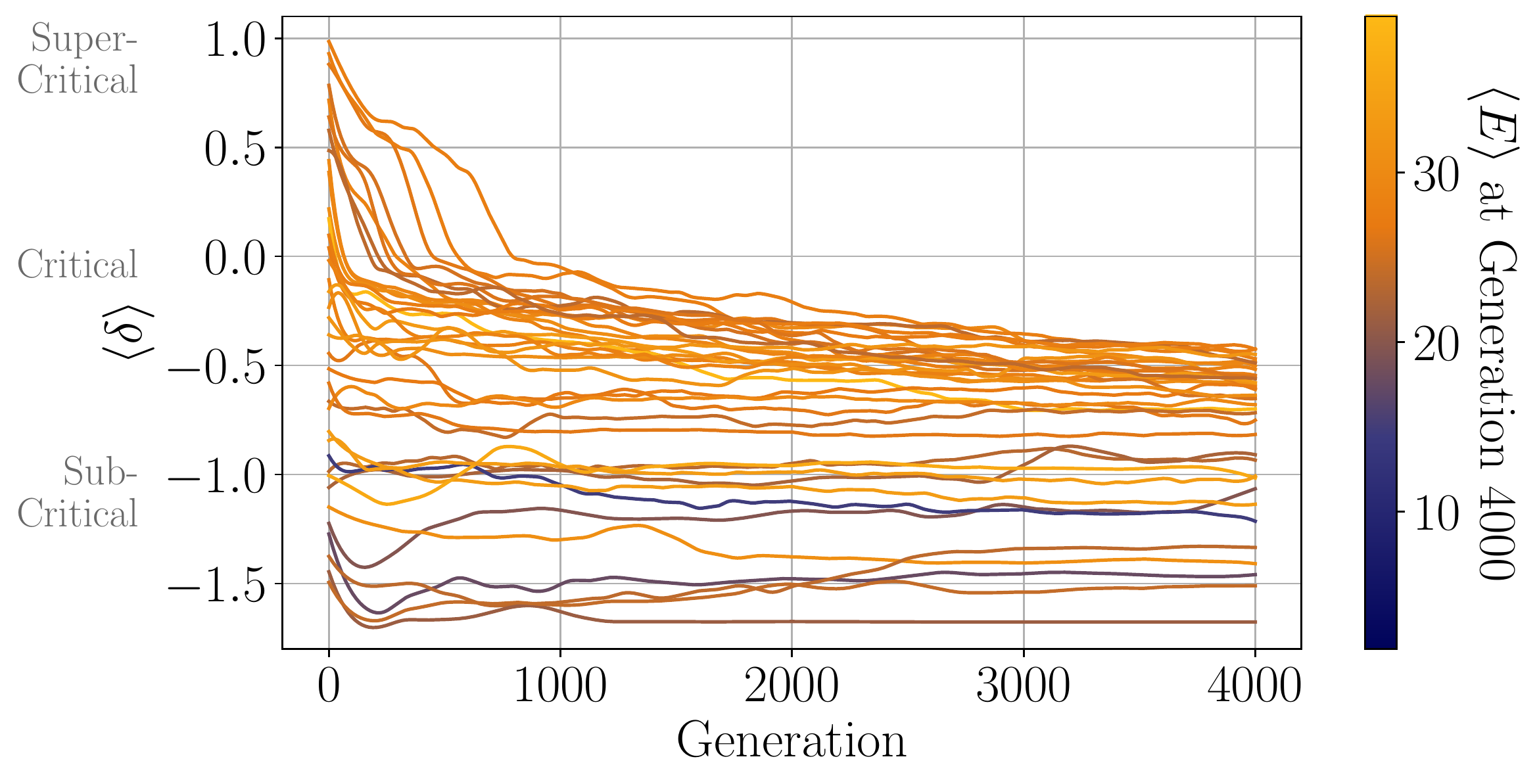} & \includegraphics[height=1.8in]{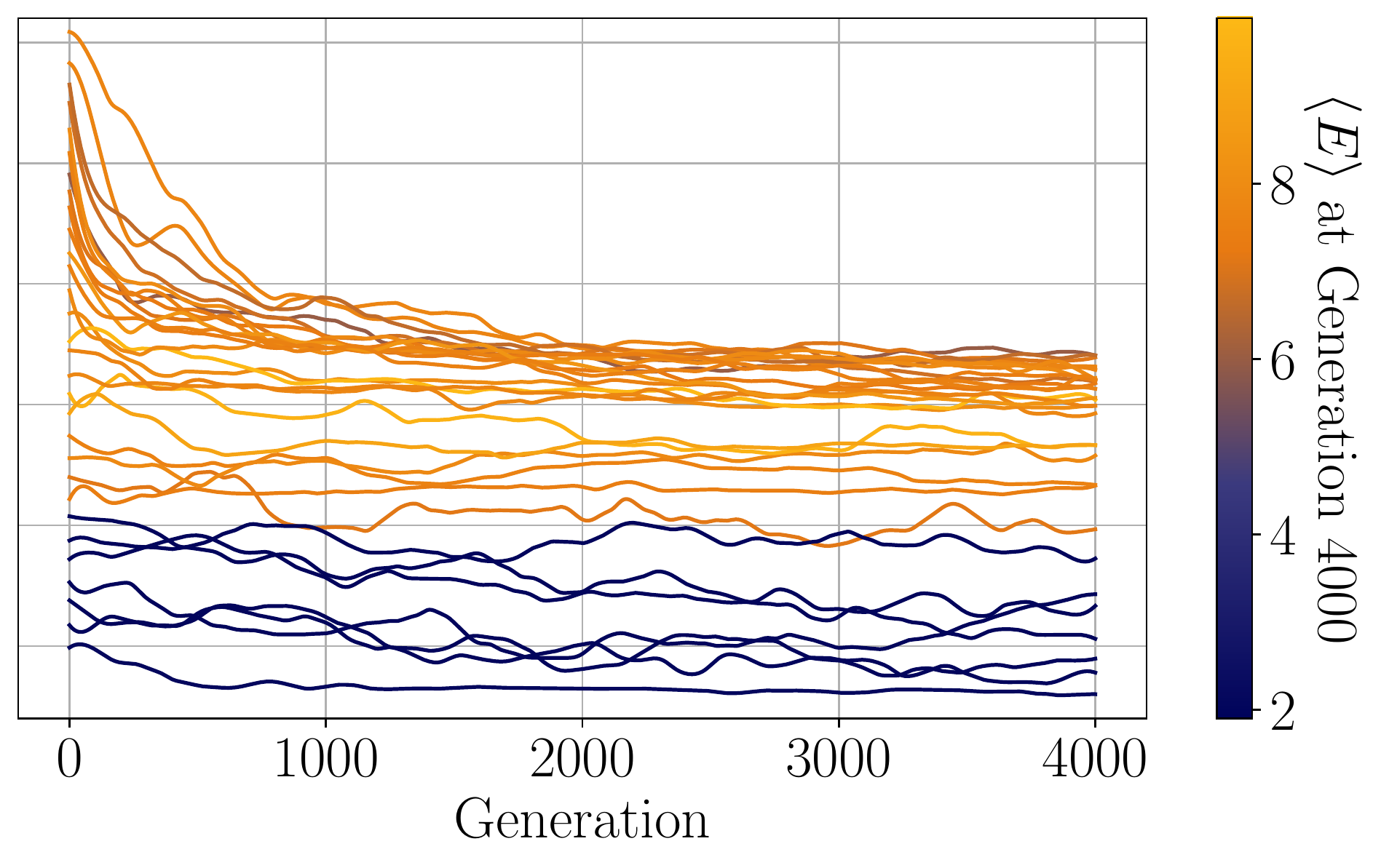}
\end{tabular}\vspace{-.5em}
\caption{Changes in the distance to criticality over the course of evolution.
Populations initiated at various distances to criticality ($\delta$ between -1.5 and 1) and evolved on a \textbf{A:} simple task and \textbf{B:} hard task.
The color indicates their fitness at generation 4000. Populations with $\delta < -1$ remain at fitness 2 for the hard task, signifying no evolutionary progress.
%\figdetail{Left plot: between 1 and -1.5 there are 38 simulations Right plot: there are 30 simulations between 1 and -1.5. For smoothing we use a Savitzky-Golay-Filter (for all figures) --> polynomial interpolation important parameters: Polynomial order (set to 3) and window length (we used different length depending on sampling rate)}
}
\label{fig:delta_vs_generation}
\end{figure*}

\begin{figure}
\centering
%width 2.3in
\includegraphics[width= 0.95\columnwidth]{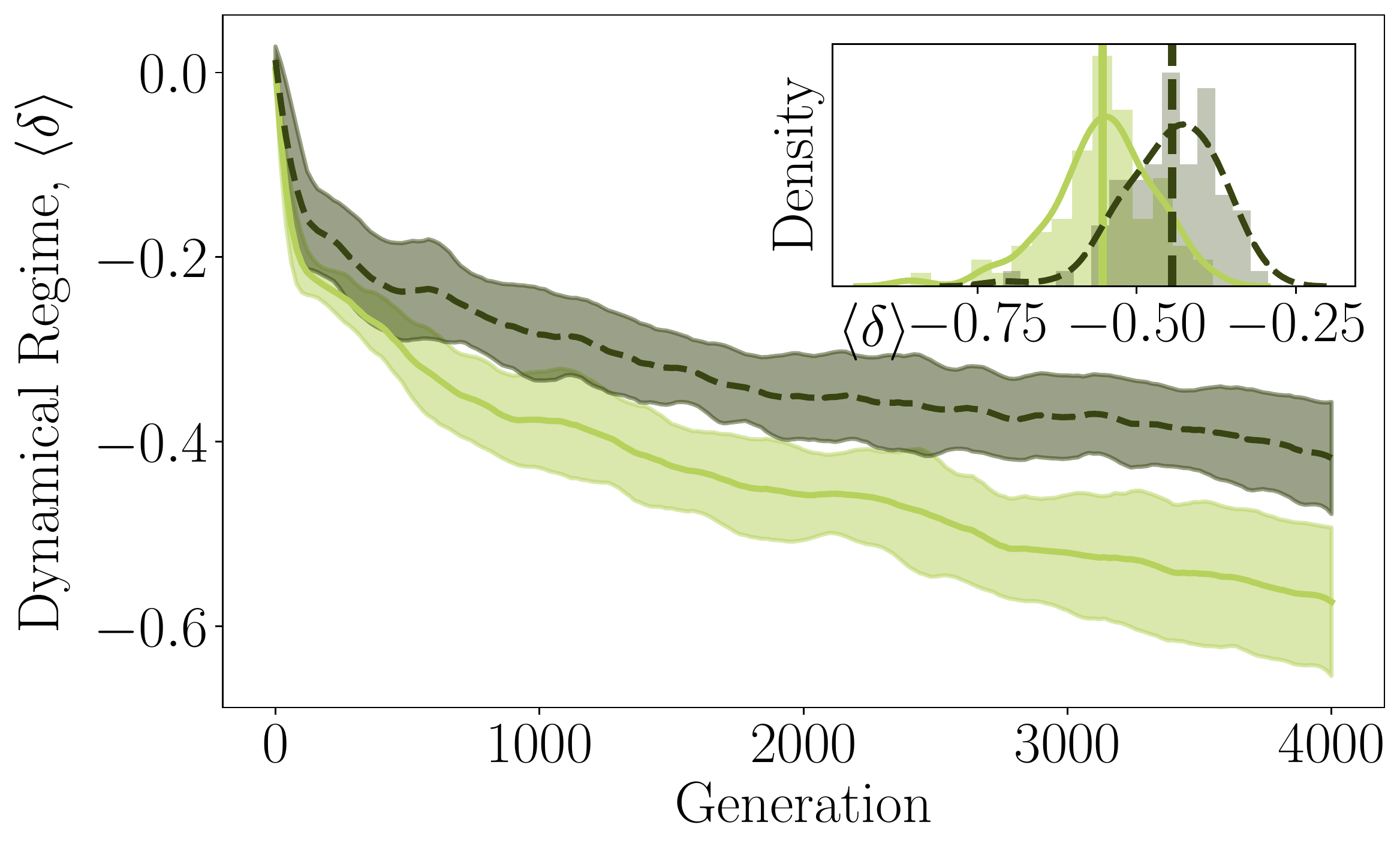}\vspace{-.5em}
\caption{The dynamical regime of the initially critical population in the harder task remains closer to the critical regime, than in the simple task throughout evolution.
Main plot: For each condition 10 populations are initiated in the critical regime end evolve for 4000 generations (dark - hard task, light - simple task).
Inset: histograms of dynamical regimes at generation 4000 of 75 populations per task, vertical lines denote means, $p < 10^{-13}$. Difference between means in inset and main panel are due to the small sample size for the main panel.
%\figdetail{2x10 simulations in the generation series plot, 2x75 simulations in the inset}
}
\label{fig:simple_vs_hard_task}
\end{figure}

For successfully evolvable populations, moderate changes in the complexity of the control network should not destroy the ability of the EA to reach a good fitness.

We test the differences in evolvability for initially critical and subcritical populations by changing the size of the network from 12 to 28 neurons.
Same as for smaller networks, the initially critical populations rapidly evolve in all realizations.
By generation 4000 they even reach a slightly larger fitness than populations of smaller networks that had evolved for the same amount of time (Figure~\ref{fig:fitness_plots_vs_generation}B).
At the same time, subcritially-initialized populations do not reach even half of their normal fitness.
We observe the same difference between the dynamical regimes when we increase the task's complexity requiring organisms to slow down almost to zero velocity in order to consume food particles (Figure~\ref{fig:fitness_plots_vs_generation}C).
In this harder task, the evolved populations' maximal fitness is expected to be lower than for the simple task.
For the initially critical populations, we still observe the same hill climbing dynamics. However, the initially subcritical populations stay at an energy level of exactly two.
This signifies that they do not use the originally supplied energy for moving and remain static throughout all 4000 generations, trapped in a local optima.

Overall, we see that although in simple tasks all populations can converge to approximately the same fitness, there exists a significant difference between the initially subcritical and initially critical populations.
Namely, a convergence of the EA for critical populations resembles hill climbing. It is stable (all populations follow very similar fitness growth) and behaves similarly for larger networks or more complex tasks.
At the same time, for subcritical populations, the EA resembles random search, which is stochastic and fails to find solutions in high-dimensional cases or for more complex tasks.

\subsection{Evolution of the dynamical regime}
\begin{figure}
\centering
%width 2.3in
\includegraphics[width= 0.95\columnwidth]{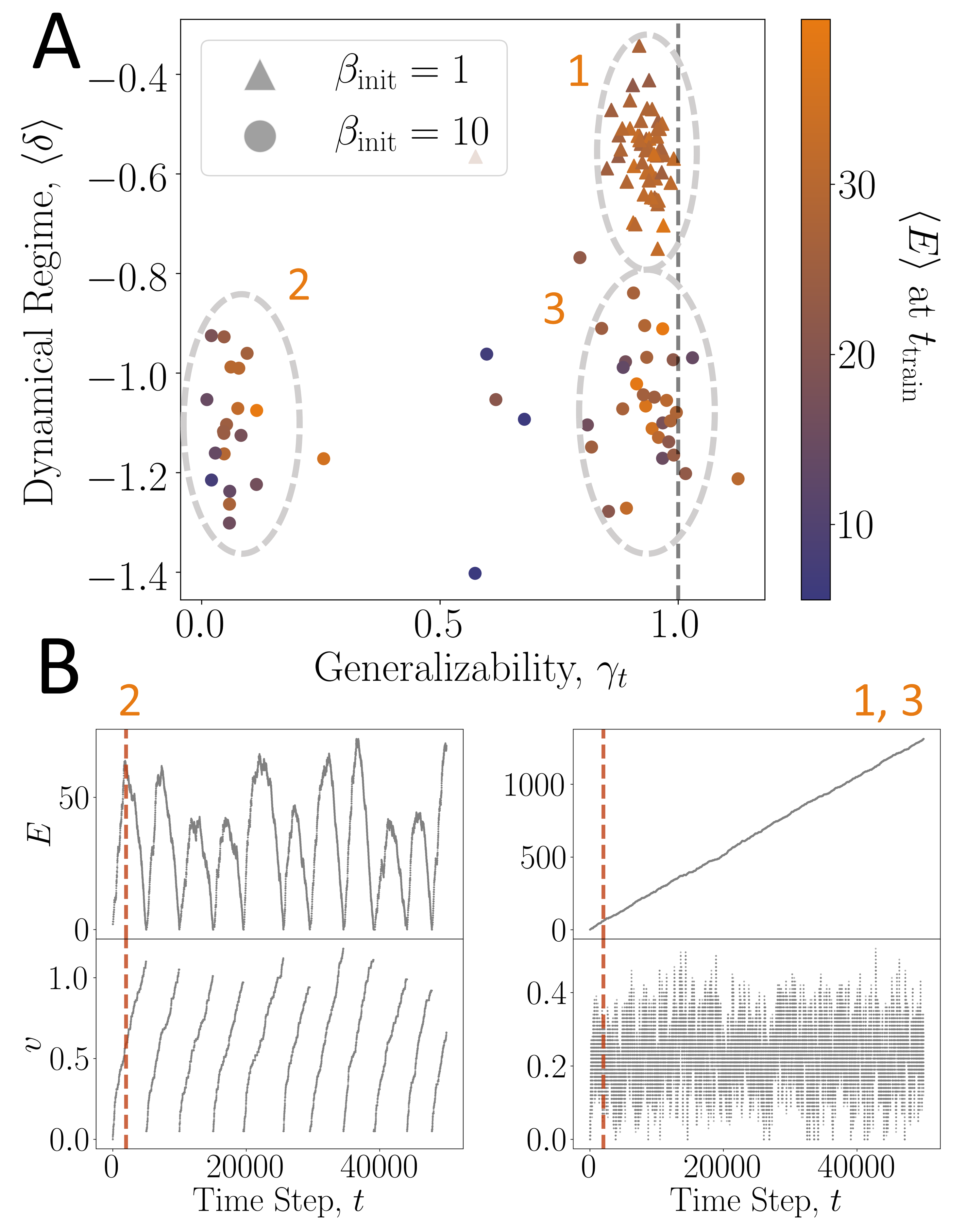}
\vspace{-.5em}
\caption{Populations initialized at criticality are always able to find solutions whose behaviors generalize in time beyond their training condition, whereas subcritical populations often overfit to the same conditions.
{\bf A:} 54 populations of each type (triangles - initially critical, circles - initially subcritical).
After 4000 generations of evolution all critical populations reach a high fitness (indicated by color) and nearly perfect (with one exception) generalizability $\gamma_t$ (Eq.~\ref{eq:generalizability_param}).
The initially subcritical organisms split into badly generalizing cluster 2 (19 populations), generalizable cluster 3 (28 populations), and 7 populations that we did not assign to any cluster. This split is independent of their attained fitness.
{\bf B:} Energy and velocity as a function of time for representative examples of the organisms from cluster 2 and clusters 1,3 (marked in panel A). Organisms in cluster 2 reach their maximal fitness (sometimes higher than in clusters 1 and 3) at the end of their lives.
The dashed orange line denotes the training lifetime of 2000 time steps.}
\label{fig:generalizability}
\end{figure}

\fs{Next, we investigate how evolution changes the state of the populations.}
To do so, we select a wide range of initial dynamical regimes ($\delta \in [-1.5, 1]$) and examine how the dynamics of populations initialized in each of these regimes change throughout evolution.

Regardless of their initial dynamical regime, all populations converge to the subcritical regime, however, with different distances from the critical point (Figure~\ref{fig:delta_vs_generation}).
We also observe that strongly subcritical populations ($\delta < -1 $) generally achieve lower fitness levels in the simple task and are unable to solve the hard task.

Both near-critical and supercritical populations rapidly change their dynamical regimes and by generation 4000 reach an intermediately subcritical regime $\delta^*$.
Strongly subcritical populations with $\delta\ll\delta^* $ remain at their initial regimes, demonstrating a lack of evolutionary mobility, whereas subcritical populations initialized at different $0 \geq \delta \geq \delta^*$  can still approach $\delta^*$ (Figure~\ref{fig:delta_vs_generation}).

Task complexity changes the dynamical regime found in the evolutionary limit. The hard task requires a smaller distance from criticality.
We check the evolution of the dynamical regime in both simple and hard tasks (Figure~\ref{fig:simple_vs_hard_task}).
We utilize the observation that all populations with an initial regime $\delta > \delta^*$ converge to similar values. Thus, we consider only initially critical populations.
We obtain the distribution of dynamical states by considering 75 independent runs of evolution in both tasks after 4000 generations.
The harder task results in a smaller distance from the critical regime (large $\delta$), and the difference is significant ($p<10^{-13}$).
The difference is observed throughout the evolution (Figure~\ref{fig:simple_vs_hard_task}, the main panel based on ten populations).

\fs{Initiating the evolution close to the critical regime is important for unknown task complexity.}
We observe that the dynamical regime never changes towards larger values, but the subcritical convergence point can be at different distances from criticality.
Thus, only starting at the critical point guarantees that the optimal dynamical state can be reached by evolution.

%Populations with different initializations also differ in their evolutionary dynamics with respect to their fitness for the simulated environment. The initially critical population rapidly adapts to the environment approaching its peak fitness relatively early on during the simulation. This quick adaptation happens consistently across many runs of the initially critical population (low variance). The initially supercitical population remains at a low fitness level until its dynamics begin approaching criticality, at which point its performance rapidly improves, resembling the initially critical population. Finally, the initially subcritical population is characterized by high variance and abrupt jumps in fitness during its evolution. Some initially subcritical organisms reach very high levels of fitness while others remain low-performing throughout the simulation.

%

\subsection{Generalizability}
\fs{ For successful biological systems robustness towards environmental change is the paramount feature, therefore, it should be used to distinguish the successfully evolved artificial organisms.}
We propose a simple measure to investigate how the model behaves outside of its explicit training conditions.
Specifically, for a population trained for the organism's lifetime $t_{\mathrm{train}}$ we define generalizability as the speed of growth of the average fitness  if the organism's lifetime is extended to $t_{\mathrm{extend}}$ . Formally:
\begin{eqnarray}
\gamma_t = \frac{\langle E_{t_\mathrm{train}} \rangle / t_\mathrm{train}}{\langle E_{t_\mathrm{extend}} \rangle / t_\mathrm{extend}}.
\label{eq:generalizability_param}
\end{eqnarray}
The \emph{stable generalizability}, $\gamma_t = 1$ corresponds to linear growth whereas sublinear behavior $\gamma_t \ll 1$  indicates possible overfitting to the particular organism's lifetime $t_{\mathrm{train}}$.

\fs{We consider initially critical ($\delta\approx0$) and initially subcritical ($\delta\approx-1$) populations evolved for 4000 generations and then test their performance for an extended lifetime of 50\,000 time steps (instead of the 2000 in training).}
 As reported in previous sections, the critical populations converge to $\delta \approx -0.55$, and they all have a similar fitness after training.
Interestingly, when increasing the organisms' lifetime, the fitness of the critical population continues to grow linearly, signifying almost perfect generalizability.
About half of the subcritical populations reach the same fitness level.
However, the subcritical populations split up into two clusters: one with generalizability close to 0 and another with generalizability close to 1 (Figure~\ref{fig:generalizability}A).
Surprisingly, there is no difference in fitness between these two clusters.
When we look more precisely at the individual organisms from the two clusters, we discover that the strategy of the organisms in the non-generalizable cluster is to increase the velocity permanently until the end of their training lifetime (Figure~\ref{fig:generalizability}B).
However, moving with such a high velocity is not compatible with the energy influx from feeding, and they break down shortly after the end of their training lifetime, this demonstrates that these organisms overfit the training conditions.
For all generalizable populations (clusters 1 and 3), the velocity fluctuates around a stable level, and the energy of the organisms grows linearly.

\fs{Overall, the initialization in the critical regime results in almost perfect generalizability of evolved populations, whereas initially strongly subcritical popuations risk overfit their training conditions.}

%A population that, in the test run, can maintain the same (or larger) surplus energy accumulation rate it exhibited in its generation 2000 training run is considered generalizable and will have a $\gamma_\mathrm{t} \approx 1$.

\subsection{Effect of genetic perturbations on the fitness}
\begin{figure}\centering
\includegraphics[width= 0.95\columnwidth]{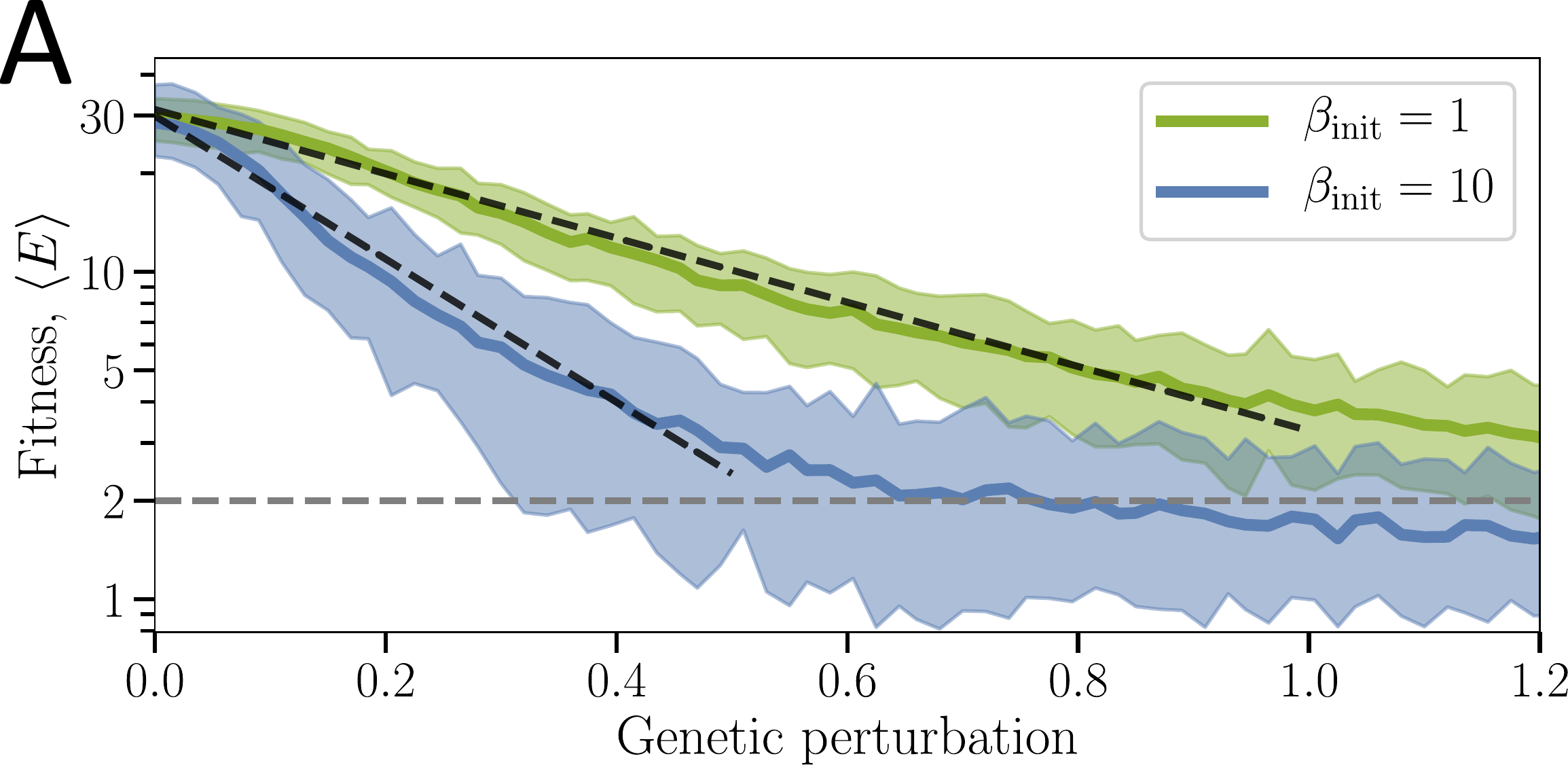}
\includegraphics[width= 0.95\columnwidth]{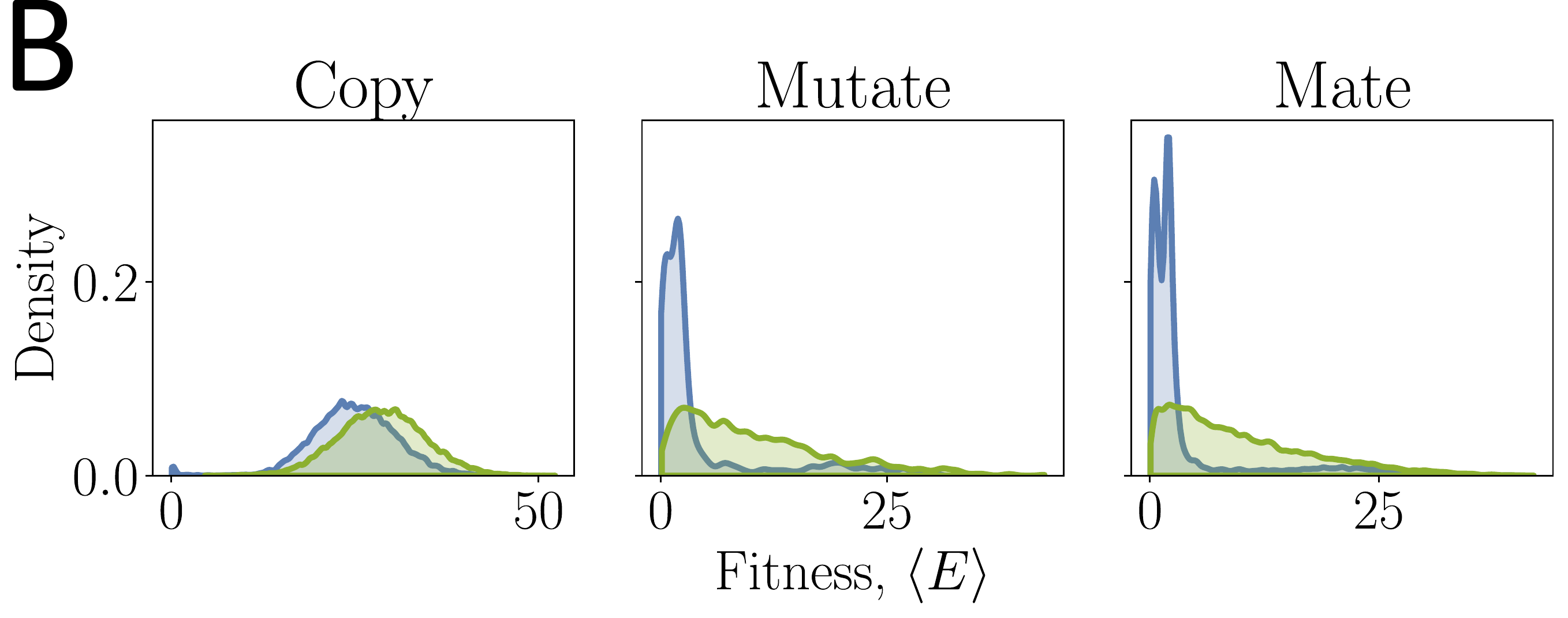}
\caption{Initially critical populations show larger genotypic stability than initially subcritical ones.
\textbf{A:} The phenotype (fitness) as a function of the genotypic perturbation (changes in connectivity) for two initial conditions: $\beta_{\mathrm{init}} = 1$ and $\beta_{\mathrm{init}} = 10$.
Dashed black lines indicate exponential fit with exponent -2.26 (critical) and -5.03 (subcritical).
%It can be seen that the dynamical regime an organism is initialized in plays an important role in characterizing its sensitivity to genetic perturbations.
\textbf{B:} The histograms of the fitness values for nearly fully evolve agents (born between generation 3500 and 4000), categorized according to the last evolutionary operator (copy, mutate, or mate) that was applied to them.
%Again, a clear difference can be seen between the two populations, where t
The $\beta_{\mathrm{init}} = 10$ agents are much less likely to remain fit when their genotype is changed by either mutation or mating. %\figdetail{\textbf{A:}perturbation constant $c_\mathrm{pert}=0.005$ (gets multiplied with perturbation factor $f_\mathrm{pert}$ on x-axis. We changed all edges $\pm c_\mathrm{pert} \cdot f_\mathrm{pert}$. Data from several populations, only copied organisms displayed. Each population is evaluated 5 times for each perturbation factor $f_\mathrm{perturb}$ \textbf{B}: Data from the last 500 generations of two handpicked simulations with similar fitness is plotted}
}
\label{fig:phenoty_genotype}
\end{figure}

\fs{Next, we examine the stability of the evolved organisms to genetic perturbations.}
We apply genetic perturbations of different magnitudes to the evolved organisms of initially critical $\beta_{\mathrm{init}}=1$ and subcritical $\beta_{\mathrm{init}}$ populations.
We perturb all weights of the connectivity matrix by randomly adding or subtracting a number $f_\mathrm{pert}$ and then evaluate the fitness of the resulting organism.
We find that fitness rapidly declines with perturbation magnitude for both populations, however the subcritical ones decline faster (Figure~\ref{fig:phenoty_genotype}A).
We evaluate the fitness decline by the slope of an exponential function fitted to the fitness.
%Furthermore, as the networks transition into subcriticality, the exponential rate for their distance from criticality in the hard and simple task are -2.26 and -5.03, respectively (Figure \ref{fig:phenoty_genotype}).
For the hard task it is $\alpha = -2.26$ and for the simple task it is $\alpha = -5.03$, which is more than double the decline rate, indicating a much higher sensitivity of subcritical systems to perturbations.
%We observe that the initially subcritical agents are more sensitive to perturbations as opposed to the initially critical populations, as can be seen in figure \ref{fig:phenoty_genotype} A, which maps the extend of genotypic perturbations to the resulting fitness.

\fs{The EA is a source of constant genetic perturbations that are necessary in the beginning of evolution but can become detrimental later}.
We consider the individual effect of the evolutionary operators (copy, mutate, and mate) on the resulting fitness of the organisms.
The variability of fitness for copying simply reflects the natural variability in community fitness rankings and organism behaviour.
However, both mating and mutation in fully evolved subcritical populations typically results in a fitness close to 2 -- signifying totally unfit organisms (Figure~\ref{fig:phenoty_genotype}B).
At the same time, initially critical organisms retain diverse fitness values after mutation and mating, some being close to the optimum. This indicates that the originally critical populations retain their evolvability as opposed to the rigid search performed by strongly subcritical populations.

%Consequently, a previous mutation or mating operation (Figure~\ref{fig:phenoty_genotype}B) is more likely to lead an initially subcritical agent towards highly sub-optimal (random) behavior (fitness $\leq 2$). On the other hand, in initially critical agents, such mutations lead to a reduction of fitness but are not likely to render an agent totally unfit. This is an effect of the rugged fitness landscape associated with the initially subcritical agents which makes them extremely sensitive to pertrubations as opposed to the smoother fitness landscape of the initially critical agents.

\section{Discussion}
\fs{We demonstrate that in various scenarios evolving populations of agents converge to a moderately subcritical state with the resulting deviation from criticality depending on the tasks difficulty. }
This might appear to be a contradiction to the previous studies, suggesting that operating close to criticality is optimal for natural systems \citep{mora2011biological_poised_at_critic, munoz2018colloquium, roli2018dynamical}.
However, a recent body of research showed that for simple tasks, operating at some distance to criticality might be an optimal solution for the sensitivity/stability tradeoff~\citep{hidalgo2014information,tomen2014marginally,villegas2016noise_subcritical,cramer_control_2020}.

%Many studies suggest that operating close to criticality is optimal for natural systems \citep{mora2011biological_poised_at_critic, munoz2018colloquium, roli2018dynamical}. We tested this hypothesis by evolving agents, controlled by Ising neural networks, to solve a foraging task in a $2$D environment. However, we observed that all populations, regardless of their initial dynamical regime, eventually evolve towards the subcritical regime.
We observe that the distance from criticality affects an agent's ability to solve complex tasks and to robustly evolve generalizable behavior.
Specifically, we observe that slightly subcritical populations are evolvable for different complexities of the control network and task, whereas strongly subcritical populations fail in both cases.
%closer to criticality performed better with respect to these properties.
Given that these properties are crucial for adaption in natural environments, we propose that living systems operate in the subcritical regime in close proximity to the critical point.
Moreover, we show that the optimal regime moves closer to criticality as we increase the task difficulty, which suggests that the optimal distance from criticality varies.
Those findings are confirmed by \citet{cramer_control_2020} as well as \citet{villegas2016noise_subcritical}, who showed that the optimal distance from criticality in the sub-critical regime decreases for higher task complexity or larger system size.
We further observe that populations can only become more subcritical during evolution and fail to decrease their distance to criticality even when this would have eventually led to superior behavior.
As it is \emph{a priori} unknown which distance from criticality will be optimal when evolving for a new task, starting at the critical point could be the only way for the evolutionary process to descend to the optimal regime.
However, in the long run this would require some sub-populations to always maintain closeness to criticality.
How this can be achieved for neuronal networks is a subject of vivid research (for a review, see~\citep{zeraati2020self,buendia2020feedback,kinouchi2020mechanisms}) and for the embodied Ising agents it remains open for further investigation.
Maintaining evolvability in simpler systems was, for instance, observed by switching between different rough energy landscapes \citep{wang2019evolving}.
The inhomogeneity of the environment and coevolution can also contribute to the preservation of the critical regime~\citep{hidalgo2014information}.
 Overall, the maintenance of evolvability throughout evolution is an important question beyond the embodied Ising agents studied here.

Our results extend and partly revise the earlier findings of \citet{khajehabdollahi2020sinas_paper}, that reported a superior evolvability of critical populations and an approximate convergence to criticality during evolution.
We confirm that the critical regime allows reliable evolvability, additionally, we extend our understanding by considering a set of tasks and architectures in the model.
However, our more precise procedure to infer the dynamical regime and the fine sampling of initial conditions uncover additional intricate dynamics.
Namely, that critical populations of Ising-agents converge to the subcritical regime, and the distance to criticality depends on the task complexity.
We also propose a new way to investigate capabilities of the resulting organisms by defining generalizability and genetic stability measures.
Both measures reveal the benefits of staying close to the critical state beyond a simple fitness comparison.

\subsection*{Acknowledgments}
%-----------------------------------------------------------

This work was supported by a Sofja Kovalevskaja Award
from the Alexander von Humboldt Foundation, endowed by the Federal Ministry of Education and Research. We acknowledge the  support from the BMBF through the T\"ubingen AI Center (FKZ: 01IS18039B).

\footnotesize
\bibliographystyle{apalike}
\bibliography{Bibliography} % replace by the name of your .bib file

\begin{thebibliography}{}

\bibitem[Aguilera and Bedia, 2017]{aguilera2017ising_neural_network}
Aguilera, M. and Bedia, M.~G. (2017).
\newblock Criticality as it could be: organizational invariance as
  self-organized criticality in embodied agents.
\newblock In {\em Artificial Life Conference Proceedings 14}, pages 21--28. MIT
  Press.

\bibitem[Aldana et~al., 2007]{aldana2007robustness_evolvability}
Aldana, M., Balleza, E., Kauffman, S., and Resendiz, O. (2007).
\newblock Robustness and evolvability in genetic regulatory networks.
\newblock {\em Journal of theoretical biology}, 245(3):433--448.

\bibitem[Balleza et~al., 2008]{balleza2008critical}
Balleza, E., Alvarez-Buylla, E.~R., Chaos, A., Kauffman, S., Shmulevich, I.,
  and Aldana, M. (2008).
\newblock Critical dynamics in genetic regulatory networks: examples from four
  kingdoms.
\newblock {\em PLoS One}, 3(6):e2456.

\bibitem[Buend\'ia et~al., 2020]{buendia2020feedback}
Buend\'ia, V., di~Santo, S., Bonachela, J.~A., and Mu\~noz, M.~A. (2020).
\newblock Feedback mechanisms for self-organization to the edge of a phase
  transition.
\newblock {\em Frontiers in Physics}, 8:333.

\bibitem[Cavagna et~al., 2010]{cavagna2010scale}
Cavagna, A., Cimarelli, A., Giardina, I., Parisi, G., Santagati, R., Stefanini,
  F., and Viale, M. (2010).
\newblock Scale-free correlations in starling flocks.
\newblock {\em Proceedings of the National Academy of Sciences},
  107(26):11865--11870.

\bibitem[Chat{\'e} and Mu{\~n}oz, 2014]{chate2014insect}
Chat{\'e}, H. and Mu{\~n}oz, M.~A. (2014).
\newblock Insect swarms go critical.
\newblock {\em Physics}, 7:120.

\bibitem[Cramer et~al., 2020]{cramer_control_2020}
Cramer, B., St{\"o}ckel, D., Kreft, M., Wibral, M., Schemmel, J., Meier, K.,
  and Priesemann, V. (2020).
\newblock Control of criticality and computation in spiking neuromorphic
  networks with plasticity.
\newblock {\em Nature communications}, 11(1):1--11.

\bibitem[De~Jong, 2006]{dejong2006evolutionary_unified_book}
De~Jong, K. (2006).
\newblock Evolutionary computation: A unified approach. bradford book.

\bibitem[De~Palo et~al., 2017]{depalo2017collective_cells}
De~Palo, G., Yi, D., and Endres, R.~G. (2017).
\newblock A critical-like collective state leads to long-range cell
  communication in dictyostelium discoideum aggregation.
\newblock {\em PLoS biology}, 15(4):e1002602.

\bibitem[Halley et~al., 2009]{halley2009collective_cells}
Halley, J.~D., Burden, F.~R., and Winkler, D.~A. (2009).
\newblock Stem cell decision making and critical-like exploratory networks.
\newblock {\em Stem Cell Research}, 2(3):165--177.

\bibitem[Hidalgo et~al., 2014]{hidalgo2014information}
Hidalgo, J., Grilli, J., Suweis, S., Munoz, M.~A., Banavar, J.~R., and Maritan,
  A. (2014).
\newblock Information-based fitness and the emergence of criticality in living
  systems.
\newblock {\em Proceedings of the National Academy of Sciences},
  111(28):10095--10100.

\bibitem[Kauffman and Levin, 1987]{kauffman1987waiting_times_double}
Kauffman, S. and Levin, S. (1987).
\newblock Towards a general theory of adaptive walks on rugged landscapes.
\newblock {\em Journal of theoretical Biology}, 128(1):11--45.

\bibitem[Kauffman, 1993]{kauffman1993origins}
Kauffman, S.~A. (1993).
\newblock {\em The origins of order: Self-organization and selection in
  evolution}.
\newblock Oxford University Press, USA.

\bibitem[Khajehabdollahi and Witkowski, 2020]{khajehabdollahi2020sinas_paper}
Khajehabdollahi, S. and Witkowski, O. (2020).
\newblock Evolution towards criticality in ising neural agents.
\newblock {\em Artificial Life}, 26(1):112--129.

\bibitem[Kinouchi et~al., 2020]{kinouchi2020mechanisms}
Kinouchi, O., Pazzini, R., and Copelli, M. (2020).
\newblock Mechanisms of self-organized quasicriticality in neuronal network
  models.
\newblock {\em Frontiers in Physics}, 8:530.

\bibitem[Mora and Bialek, 2011]{mora2011biological_poised_at_critic}
Mora, T. and Bialek, W. (2011).
\newblock Are biological systems poised at criticality?
\newblock {\em Journal of Statistical Physics}, 144(2):268--302.

\bibitem[Munoz, 2018]{munoz2018colloquium}
Munoz, M.~A. (2018).
\newblock Colloquium: Criticality and dynamical scaling in living systems.
\newblock {\em Reviews of Modern Physics}, 90(3):031001.

\bibitem[R{\"a}m{\"o} et~al.,
  2007]{pauli_ramo2007_critical_information_propagation_subcritical_noise}
R{\"a}m{\"o}, P., Kauffman, S., Kesseli, J., and Yli-Harja, O. (2007).
\newblock Measures for information propagation in boolean networks.
\newblock {\em Physica D: Nonlinear Phenomena}, 227(1):100--104.

\bibitem[R{\"a}m{\"o} et~al., 2006]{ramo2006perturbation}
R{\"a}m{\"o}, P., Kesseli, J., and Yli-Harja, O. (2006).
\newblock Perturbation avalanches and criticality in gene regulatory networks.
\newblock {\em Journal of Theoretical Biology}, 242(1):164--170.

\bibitem[Roli et~al., 2018]{roli2018dynamical}
Roli, A., Villani, M., Filisetti, A., and Serra, R. (2018).
\newblock Dynamical criticality: overview and open questions.
\newblock {\em Journal of Systems Science and Complexity}, 31(3):647--663.

\bibitem[Schneidman et~al., 2006]{schneidman2006neural_cultures}
Schneidman, E., Berry, M.~J., Segev, R., and Bialek, W. (2006).
\newblock Weak pairwise correlations imply strongly correlated network states
  in a neural population.
\newblock {\em Nature}, 440(7087):1007--1012.

\bibitem[Tkacik et~al., 2015]{tkacik2015thermodynamics}
Tkacik, G., Mora, T., Marre, O., Amodei, D., Palmer, S.~E., and Bialek, W.
  (2015).
\newblock {Thermodynamics and signatures of criticality in a network of
  neurons}.
\newblock {\em PNAS}, 112(37).

\bibitem[Tomen et~al., 2014]{tomen2014marginally}
Tomen, N., Rotermund, D., and Ernst, U. (2014).
\newblock Marginally subcritical dynamics explain enhanced stimulus
  discriminability under attention.
\newblock {\em Frontiers in systems neuroscience}, 8:151.

\bibitem[Villegas et~al., 2016]{villegas2016noise_subcritical}
Villegas, P., Ruiz-Franco, J., Hidalgo, J., and Mu{\~n}oz, M.~A. (2016).
\newblock Intrinsic noise and deviations from criticality in boolean
  gene-regulatory networks.
\newblock {\em Scientific Reports}, 6:34743.

\bibitem[Wang and Dai, 2019]{wang2019evolving}
Wang, S. and Dai, L. (2019).
\newblock Evolving generalists in switching rugged landscapes.
\newblock {\em PLoS computational biology}, 15(10):e1007320.

\bibitem[Zeraati et~al., 2020]{zeraati2020self}
Zeraati, R., Priesemann, V., and Levina, A. (2020).
\newblock Self-organization toward criticality by synaptic plasticity.
\newblock {\em arXiv preprint arXiv:2010.07888}.

\bibitem[Zierenberg et~al., 2020]{zierenberg2020tailored}
Zierenberg, J., Wilting, J., Priesemann, V., and Levina, A. (2020).
\newblock Tailored ensembles of neural networks optimize sensitivity to
  stimulus statistics.
\newblock {\em Physical Review Research}, 2(1):013115.

\end{thebibliography}

\end{document}